\pdfoutput=1

\documentclass[11pt]{article}
\usepackage{natbib}
\usepackage[dvipsnames]{xcolor}
\usepackage{tcolorbox}
\usepackage{authblk}
\usepackage{ACL2025}
\usepackage{hyperref}
\usepackage{times}
\usepackage{latexsym}
\usepackage{booktabs}
\usepackage[T1]{fontenc}

\usepackage[utf8]{inputenc}

\usepackage{microtype}

\usepackage{inconsolata}
\usepackage{array} 

\usepackage{graphicx}
\usepackage{todonotes}
\usepackage{amsmath}
\usepackage{multirow}
\usepackage{multicol}
\usepackage{caption}
\usepackage{subcaption}
\usepackage{float}

\usepackage[labelfont=bf]{caption}
\usepackage{adjustbox}

\newlength{\wdth}

\title{\textit{ReVision:} A Dataset and Baseline VLM for Privacy-Preserving Task-Oriented Visual Instruction Rewriting}  

\author{
Abhijit Mishra$^\dagger$,
Mingda Li$^\dagger\ddagger$,
Hsiang Fu$^\dagger\S$,
Richard Noh$^\dagger$,
Minji Kim$^\dagger$ \\
$^\dagger$ School of Information, The University of Texas at Austin \\
$\ddagger$ Department of Statistics and Data Science, Yale University \\
$\S$ School of Computing and Augmented Intelligence, Arizona State University \\
\texttt{\{abhijitmishra, mingdali, seanfu, rjnoh, minji.kim\}@utexas.edu}
}

\begin{document}
\maketitle

\begin{abstract}
Efficient and privacy-preserving multimodal interaction is essential as AR, VR, and modern smartphones with powerful cameras become primary interfaces for human-computer communication. Existing powerful large vision-language models (VLMs) enabling multimodal interaction often rely on cloud-based processing, raising significant concerns about (1) visual privacy by transmitting sensitive vision data to servers, and (2) their limited real-time, on-device usability. This paper explores \textit{Visual Instruction Rewriting}, a novel approach that transforms multimodal instructions into text-only commands, allowing seamless integration of lightweight on-device instruction rewriter VLMs \textbf{(250M parameters)} with existing conversational AI systems, enhancing vision data privacy. To achieve this, we present a dataset of over 39,000 examples across 15+ domains and develop a compact VLM, pretrained on image captioning datasets and fine-tuned for instruction rewriting. Experimental results, evaluated through NLG metrics such as BLEU, METEOR, and ROUGE, along with semantic parsing analysis, demonstrate that even a quantized version of the model (<500MB storage footprint) can achieve effective instruction rewriting, thus enabling privacy-focused, multimodal AI applications.
\end{abstract}

\section{Introduction}
\label{sec:intro}

The rapid integration of conversational AI into AR, VR, smartphones, and wearables has heightened the demand for multimodal systems that can interpret text, images, speech, and gestures seamlessly. Devices like the Meta Ray-Ban Smart Glasses, Apple Vision Pro, and tools like Google Lens enable users to ask specific questions about their visual surroundings - yet entire images, often containing sensitive background data unrelated to the query, are transmitted to cloud-based large or semi-large vision-language models (VLMs) \cite{chen2023pali, liu2023visual, qwen-vl}, posing serious privacy risks. This highlights a key challenge: \textit{executing task-oriented multimodal commands while preserving user privacy}. On-device processing is increasingly favored to avoid exposing private content such as faces, locations, or documents. However, while large VLMs like PaLI-X, LLaVA, and Qwen-VL excel at complex tasks, they are too resource-intensive for local use, and smaller, more private models often lack the broad world knowledge needed for robust multimodal understanding.

To address this, we propose \textit{ReVision}, an approach based on \textit{Visual Instruction Rewriting} that converts multimodal instructions into text-only commands. By transforming complex visual interactions into structured text, existing lightweight on-device conversational AI models can efficiently process user instructions without sending either sensitive visual or textual data to external servers. We introduce a curated dataset consisting of \texttt{$\langle$ image, original instruction, rewritten instruction $\rangle$} triplets, covering diverse real-world tasks. A freshly built, compact, 250-million-parameter vision-language model \cite{liu2023visual} is fine-tuned on this dataset and evaluated using NLG metrics (such as BLEU, METEOR, ROUGE) and semantic parsing accuracy.

Our findings demonstrate that our compact model achieves an acceptable level of rewriting capabilities, and performs better compared to popular baselines such as PaliGemma-v2 \cite{steiner2024paligemma} and Qwen2VL \cite{wang2024qwen2} in zero-shot settings and a fully fine-tuned version of a 250M parameter VLM (SmolVLM) \cite{marafioti2025smolvlm}. Additionally, even an 8-bit quantized version of our model (<500MB on storage disk) achieves effective instruction rewrites while maintaining a small computational footprint. We strongly believe this approach bridges the gap between large-scale multimodal AI and privacy-centric, on-device execution, ensuring secure, real-time interaction with AR/VR and smartphone interfaces.

The contributions of this paper are as follows:
\begin{itemize}
    \item A novel dataset for Visual Instruction Rewriting, covering 15+ intent domains, 1,700+ personal images, and 39,000+ examples.
    \item A 250M-parameter baseline VLM using the Perceiver Resampler~\cite{laurenccon269587869matters}, pretrained on captioning datasets and fine-tuned on our rewriting dataset.
    \item Empirical validation with NLG and semantic parsing metrics, demonstrating effectiveness using GPT-4o as an on-device parser proxy.
\end{itemize}

The Code\footnote{\url{https://github.com/abhijitmishra/visual_instruction_rewriting}}, Dataset\footnote{\url{https://huggingface.co/datasets/hsiangfu/multimodal_query_rewrites}} and Models\footnote{\url{https://huggingface.co/hsiangfu/ReVision-250M-256-16-baseline}} have been released for academic use. 
\section{Related Work}
\label{sec:related_work}
Instruction or query rewriting and semantic parsing have been widely explored in conversational AI to improve query understanding and response generation. Early methods relied on rule-based transformations and supervised learning \cite{semantic_parsing_survey}, while recent advances leverage LLMs for dynamic query refinement \cite{ye2023large, mo2023convgqr}. Generative query rewriting frameworks such as LLM-R2 \cite{llm-r2} enhance text ranking, and personalized query rewriting methods \cite{cho-etal-2021-personalized} refine queries based on user preferences. However, these techniques focus primarily on textual query transformations and do not extend to multimodal task-oriented instruction processing. Visual instruction tuning has emerged as a key development in multimodal AI, with models like LLaVA \cite{liu2023visual} and PaLI-X \cite{chen2023pali} demonstrating strong vision-language capabilities. While these models excel in multimodal question answering, they are not optimized for rewriting task-oriented instructions. Similarly, Patel et al. \cite{patel2020generating} explore generating natural questions from images for multimodal assistants, but their work focuses on question generation rather than instruction rewriting. Unlike these approaches, our work introduces a dedicated dataset and a compact model for Visual Instruction Rewriting, specifically designed to convert multimodal user instructions into structured text for privacy-preserving, on-device execution.  

The closest work to ours is MARRS \cite{ates2023marrs}, which integrates multimodal reference resolution with query rewriting to improve conversational grounding. However, MARRS relies on rule-based replacements after reference resolution in a non-VLM setting, whereas our approach focuses on learning-based instruction rewriting to enable structured task execution from multimodal inputs. Other highly relevant studies are by \newcite{zhang2022can} and \newcite{wei2021visual}, which investigate whether open-domain text-based QA systems can handle visual knowledge questions by reformulating them into purely textual queries. Their work highlights the effectiveness of query rewriting in bridging the gap between vision and language using a modular approach different from ours but aligns closely with our goal of rewriting multimodal instructions into structured text. However, while their approach focuses on adapting visual questions for open-domain QA, our work is specifically designed for task-oriented instruction execution, making it applicable to a broader set of real-world multimodal interactions.

\begin{figure*}[t]
  \centering
  \includegraphics[width=\textwidth, height = 8cm]{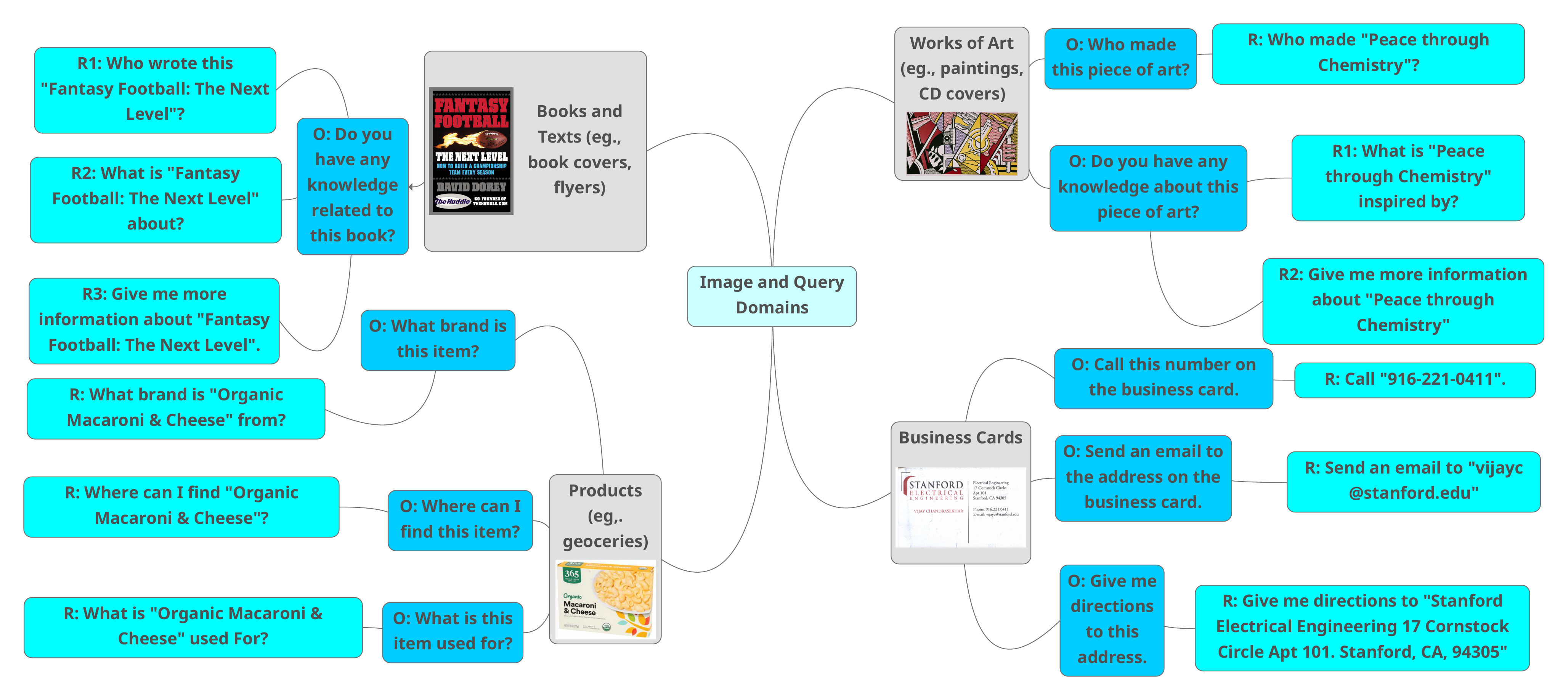} 
  \caption{Mindmap showing Data Collection and Rewrite Desiderata. O = Original Query. R = Rewritten Query.}
  \label{fig:mindmap}
\end{figure*}

\section{Constructing a Dataset for Visual Instruction Rewriting}
\label{sec:datasets}

Task-oriented conversational AI systems rely on a semantic parser to interpret user intent and extract \\structured arguments \cite{louvan2020recent, aghajanyan2020conversational}. For example, when a user says,\textit{ "Add the team meeting to my calendar for Friday at 3 PM"}, the system must parse the intent (\textit{CreateCalendarEvent}) and extract arguments such as the \textit{EventTitle} (``team meeting''), \textit{EventDate} (``Friday''), and \textit{EventTime} (``3 PM'') to schedule the event correctly. Unlike purely text-based interactions, multimodal instructions, particularly those directed at conversational AI assistants on AR/VR devices (\textit{e.g.,} Apple's Siri for Apple Vision Pro), introduce additional challenges such as ellipsis and coreference resolution. For instance, a user may look at a book cover and ask, \textit{“Who wrote this?”} or point at a product in an AR interface and say, \textit{“How much does this cost?”} Traditional text-based semantic parsers struggle with such instructions since critical visual context is missing. Thus, to bridge the gap between multimodal input and existing conversational AI stacks, we introduce a dataset specifically designed for \textit{rewriting multimodal instructions} into structured text that can be processed by standard text-based semantic parsers. Figure \ref{fig:mindmap} illustrates a representation of the dataset collection requirement, highlighting the transformation of multimodal inputs into text-based rewrites.

To construct our dataset, we first define an ontology of intents and arguments, as existing ontologies in conversational AI and semantic parsing are often proprietary and unavailable for research use. We take inspiration from \newcite{goel2023presto} for ontology and extend it to accommodate multimodal task-oriented interactions. Figure \ref{fig:intent_argument_box} presents an overview of the intents and arguments in our ontology. Next, we curate a diverse set of images covering various real-world multimodal interaction scenarios, including book covers, product packaging, paintings, mobile screenshots, flyers, signboards, and landmarks. These images are sourced from publicly available academic datasets, such as OCR-VQA\footnote{\url{https://ocr-vqa.github.io/}}, CD and book cover datasets, Stanford mobile image datasets\footnote{\url{http://web.cs.wpi.edu/~claypool/mmsys-dataset/2011/stanford/}}, flyer OCR datasets\footnote{\url{github.com/Skeletonboi/ocr-nlp-flyer.git}}, signboard classification datasets\footnote{\url{github.com/madrugado/signboard-classification-dataset}}, Google Landmarks\footnote{\url{github.com/cvdfoundation/google-landmark}}, and Products-10K\footnote{\url{https://products-10k.github.io/}}.

\begin{table}[t]
    \centering
    \scriptsize
    \label{tab:dataset_statistics}
    \begin{tabular}{llccc}
        \toprule
        \textbf{Category} & \textbf{Total} & \textbf{Train} & \textbf{Test} \\
        \midrule
        Book              & 485 / 500                               & 386 / 399                               & 101 / 101                               \\
        Business Card     & 26 / 960                                & 26 / 772                                & 26 / 188                                \\
        CD               & 27 / 1,020                              & 27 / 835                                & 27 / 185                                \\
        Flyer & 159 / 5,940                             & 159 / 4,742                             & 159 / 1,198                             \\
        Landmark         & 511 / 19,274                            & 511 / 15,420                            & 511 / 3,854                             \\
        Painting & 27 / 980                                & 27 / 774                                & 27 / 206                                \\
        Product          & 499 / 10,349                            & 499 / 8,276                             & 492 / 2,073                             \\
        \midrule
        \textbf{Total}   & \textbf{1,734 / 39,023}                 & \textbf{1,635 / 31,218}                 & \textbf{1,343 / 7,805}                  \\
        \bottomrule
    \end{tabular}
    \caption{Number of Images/Instructions per Category}
    \label{tab:sources}
\end{table}

\begin{table}[t]
    \centering
    \footnotesize
    \begin{tabular}{l  c}
        \toprule
         \textbf{Annotator}& \textbf{Percentage of Correct Captions}\\ 
         \midrule
         Annotator 1	& 90.62\%\\ 
         Annotator 2	& 87.23\%\\
         Annotator 3	& 86.35\%\\
         \midrule
         \textbf{At least two }& \textbf{92.18}\%\\
         \midrule
         \textit{All three }& \textit{74.63}\% \\
         \bottomrule
    \end{tabular}
    \caption{GPT-4 Instruction Rewriting Validation Results from Amazon Mechanical Turk }
    \label{tab:annotator_data}
\end{table}
\begin{figure}[t]
  \includegraphics[width=\columnwidth]{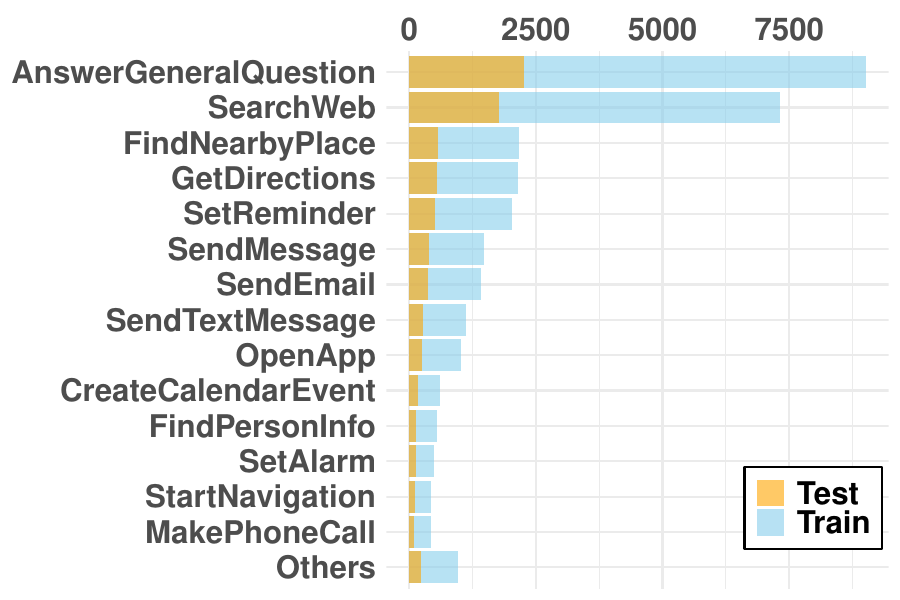}
  \caption{Dataset Distributions By Intent}

  \label{fig:intent}
\end{figure}

Upon identifying and verifying the images, we employ the GPT-4 model from OpenAI \cite{achiam2023gpt} to systematically generate and refine multimodal instructions into rewritten text-based instructions. The process begins with a bootstrap phase, where GPT-4 is prompted to generate 20 direct questions per image by explicitly referencing visible objects or textual elements while adhering to the intent list defined in Figure \ref{fig:intent_argument_box}. A second prompting phase then validates the generated questions against the corresponding image, filtering out ambiguous or irrelevant instructions to ensure alignment with the visual context. 

In the rewriting phase, GPT-4 is tasked with paraphrasing the validated instructions, ensuring that the transformed questions are fully self-contained and interpretable without requiring the image. This transformation is crucial for enabling multimodal conversational AI systems to process instructions using purely text-based stacks. Finally, a verification phase prompts the model to assess the rewritten questions in relation to both the original instruction and the image, ensuring semantic fidelity and eliminating inconsistencies. This multi-stage prompting strategy resulted in a dataset of \textbf{39,023} original-rewritten instruction pairs, derived from \textbf{1,734} images, with an 80\%-20\% train-test split. Table \ref{tab:sources} provides a breakdown of image sources.

While automated validation ensures consistency across different stages, human evaluation remains critical for verifying the dataset’s reliability. To this end, we conducted an annotation task via Amazon Mechanical Turk (AMT) to validate rewritten instructions within the test set for indirect image-based instructions. Each annotation task followed a structured validation guideline, where annotators reviewed an image, its original multimodal instruction, and the rewritten text-only instruction, determining whether the reformulation preserved the intent and meaning of the original instruction. Annotators were instructed to select "Accept" if the rewritten instruction was correct or "Reject" if it failed to capture the original meaning. Annotators are incentivized appropriately for this binary grading task. Agreement analysis, as shown in Table \ref{tab:annotator_data}, indicates that in 92.2\% of cases, at least two annotators agreed on "Accept," while 74.6\% of instructions achieved full consensus across all three annotators. Despite a Fleiss' Kappa score of 0.278—suggesting fair inter-annotator agreement—the high rate of majority consensus supports the dataset’s reliability for real-world use. Given these results, we publicly release the full dataset along with raw AMT responses, enabling further analysis, filtering, and refinements by the research community.

Figure \ref{fig:intent} presents the distribution of intents in our dataset, categorized into training and test splits. The distribution reflects practical usage patterns in real-world multimodal conversational AI systems, with a higher occurrence of general QA and web search, alongside diverse task-oriented intents such as reminders, messaging, and navigation, ensuring coverage of frequent user interactions.
\section{Developing Small-Scale VLM for Visual Instruction Rewriting}
\begin{figure}[t]
  \centering
  \includegraphics[width=\columnwidth]{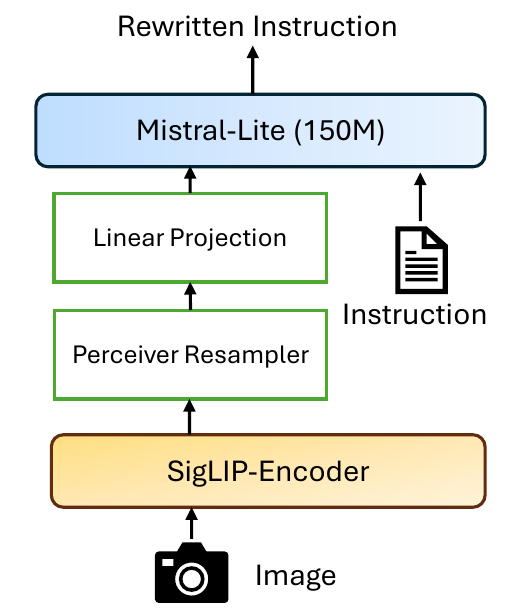}
  \caption{Revision Model Architecture}
  \label{fig:diagram}
\end{figure}

We develop a lightweight vision-language model (shown in Figure \ref{fig:diagram}) tailored for instruction-following tasks by integrating a pretrained vision encoder with an instruction-tuned language model, following the popular multimodal fusion approach \cite{zhang2024vision}. Since vision encoders and instruction-tuned language models operate in different embedding spaces, a multimodal projector \cite{liu2023visual} is used to align the encoded image features with the token embedding space of the language model. Our approach is similar to PaLI-Gemma \cite{beyer2024paligemma}, where an image encoder based on the SigLIP architecture \cite{zhai2023sigmoid} extracts $D$-dimensional image encodings for $N$ patches from a single input image, say $V_1, V_2, ..., V_N$).  Building on \newcite{laurenccon269587869matters}, who demonstrated that using a sampling technique to extract the most relevant $M$ patch encodings from a larger set of $N$ samples improves efficiency, we employ \textit{Perceiver Sampler} \cite{jaegle2021perceiver} to downsample the $N$ patch embeddings into $M$ D-dimensional encodings. These image encodings are then mapped into a shared embedding space via a linear multimodal projector, ensuring compatibility with the language model’s $H$-dimensional token embeddings. We fix $K$ at 64. The projected image embeddings ($H_1, H_2, ..., H_M$) are concatenated with the token embeddings extracted from the tokenized textual input ($H_1, H_2, ..., H_K$), where $K$ represents the number of input tokens. The combined embeddings are then processed by the language model to generate responses. To ensure consistency in input representation, we apply image preprocessing, tokenization, and chat template formatting, making the model familiar with structured multimodal input formats. 

Although large-scale vision-language models typically involve hundreds of millions of parameters, our focus is on designing a compact and efficient model capable of running on-device. To maintain a parameter budget under 250M, we select a small SigLIP encoder \cite{zhai2023sigmoid} (\texttt{google/siglip-base-patch16-256}), which processes images of size $256 \times 256$ by dividing them into $16 \times 16$ patches, with 768 dimensions in hidden layers. The language model is a 150M-parameter instruction-tuned model from OuteAI\footnote{\url{https://www.outeai.com}} (\texttt{OuteAI/\allowbreak Lite-Mistral-150M-\allowbreak v2-Instruct}) based on the Mistral architecture \cite{jiang2023mistral}, featuring a vocabulary size of 32,768 and a hidden dimension of 768. Since the hidden dimensions of both the vision encoder and the language model are identical, the projector acts purely as a dimensional transformer without altering the shape of the embeddings. While the model's limited size may impact its ability to handle multi-turn conversations, it is well-suited for single-turn multimodal instruction rewriting tasks. Additionally, since the model is designed for multimodal deixis resolution, it may not be effective for resolving text-only references in extended conversations\cite{ates2023marrs}.  

\subsection{Model Pretraining}
To pretrain the model, we adopt an end-to-end training strategy, leveraging datasets from three key sources: (a) LLaVA-ReCap-CC3M, (b) LLaVA-Pretrain, and (c) LLaVA-CC3M-Pretrain-595K. These datasets are curated from large-scale image-text corpora, including LAION \cite{schuhmann2021laion400mopendatasetclipfiltered}, Conceptual Captions (CC) \cite{sharma-etal-2018-conceptual}, and SBU \cite{NIPS2011_5dd9db5e}, which are filtered for balanced concept distribution and enhanced with synthetic captions generated via BLIP to improve vision-language alignment \cite{lmms2023llava, liu2023pretrain, liu2023cc3m}. Specifically, LLaVA-ReCap-CC3M focuses on re-captioning images to improve concept coverage, while LLaVA-Pretrain consists of 558K image-caption pairs, forming a strong foundational dataset for multimodal alignment. The LLaVA-CC3M-Pretrain-595K dataset, derived from Conceptual Captions 3M, provides a rich set of image-text pairs to enhance model robustness. The total number of examples is thus a little more than 4M. Despite some redundancy in images across datasets, we ensure sufficient data diversity and scale to instill basic image-text alignment capabilities in our pretrained model.  

For pretraining, we use the following configurations: a batch size of 16, trained for 2 epochs, using the AdamW optimizer with a learning rate of $2\times10^{-5}$ and a linear learning rate schedule. The training was conducted on consumer-grade GPUs (NVIDIA RTX 3090) over 3 days, using PyTorch and Hugging Face’s Transformers library for implementation. We refer to our pretrained model as \texttt{ReVision-250M-64-16}.

\subsection{Model Fine-Tuning}
For the instruction rewriting task, we conduct fine-tuning under multiple configurations, trained on our dataset (\ref{sec:datasets}). We will refer to the rewritten prompts from this dataset as the ``reference'' prompts. Below, we describe the fine-tuning setups and the rationale behind integrating metadata-driven enhancements to improve performance on text-dense images.

\begin{itemize}
\item \textbf{ReVision-BL}: This is the baseline fine-tuned model. The input consists of an image, a rewrite prompt, and an instruction, while the model generates a rewritten version of the instruction in response.

\item \textbf{ReVision-Metadata}: In this, we augment the input with ``metadata'', namely the \textit{image caption} and \textit{an external OCR-extracted text}. To differentiate the rewrite prompt and instruction from the auxiliary metadata, we prefix the prompt and metadata sections with \texttt{<task>} and \texttt{<data>}, respectively. Collectively, the input consists of an image, a prefixed rewrite prompt and instruction, and a prefixed caption and OCR text and the output is a rewritten instruction. 
\end{itemize}

The motivation for integrating metadata arises from the limitations of small-scale vision-language models (VLMs). Despite being optimized for rewriting tasks, small VLMs struggle with extracting embedded text from images. OCR is a specialized capability distinct from traditional vision-language alignment \cite{lamm2024can, nagaonkar2025benchmarking}. However, most modern devices are equipped with built-in OCR and image description capabilities, making it practical to supplement the model with external text recognition systems. To systematically evaluate this approach, we present two different metadata extraction:

\begin{itemize}
\item \textbf{GPT-4o\_Caption+OCR}: We use GPT-4o to generate both captions and OCR-extracted text, simulating a practical situation where a device is usually equipped with an advanced OCR and captioning system.
\item \textbf{Self\_Caption+EasyOCR}: We use rewriter models to generate captions themselves using the simple prompt: \textit{“Caption this:”}. For OCR, we employ EasyOCR\footnote{\url{https://github.com/JaidedAI/EasyOCR}}, a lightweight text extraction model based on the CRAFT algorithm \cite{baek2019character}, simulating a low-resource on-device setting.
\end{itemize}

The fine-tuning procedure follows a similar framework to pretraining but with optimized hyperparameters for smaller-scale adaptation. The vision encoder is frozen during fine-tuning, and the number of training epochs is increased from 2 to 5 to compensate for the smaller dataset size. The batch size remains at 16, but gradient accumulation steps are reduced from 4 to 1, allowing for more frequent model updates. The learning rate remains stable at $2\times10^{-5}$ with the same linear rate schedule, maintaining a conservative optimization approach. Additionally, the number of warm-up steps is lowered from 100 to 10, reflecting the shorter training duration. To simulate a realistic fine-tuning environment where such models could be updated on-device, we conduct fine-tuning on a consumer-grade desktop equipped with an NVIDIA GeForce RTX 2070 SUPER (8GB VRAM). Each fine-tuning run took approximately 5.5 to 6 hours.

For baseline comparisons, we evaluate our model against state-of-the-art small-scale VLMs: PaliGemma-v2 (10B) and QwenVL-v2 (7B), known for strong performance in OCR, captioning, and multimodal reasoning. However, deploying these models on-device is impractical without high-end GPUs. To ensure a fair comparison, we assess them as-is with optimized prompting but without fine-tuning, reflecting real-world constraints. While fine-tuning could improve accuracy, their size and hardware demands make them unsuitable for mobile applications, thus highlighting the need for lightweight models like ours. 

For deployable small VLM baselines, we include \textbf{Smol-VLM} (256M) \cite{marafioti2025smolvlm} - the smallest publicly available off-the-shelf VLM\footnote{\url{https://huggingface.co/HuggingFaceTB/SmolVLM-Instruct}} to date. We fine-tuned it on our dataset using the same configuration as our primary model, observing steady loss reduction and convergence. 

To clarify the distinction between ReVision and a simple captioning + text fusion approach, and to assess the impact of our dataset, we also compare against two \textbf{TextOnly} baselines: (a) Qwen2.5-0.5B \cite{qwen2.5}, and (b) Mistral-Lite (our custom text backbone), both fine-tuned in a pure text-to-text setting with instructions, EasyOCR outputs, and GPT-4-generated image captions. These comparisons help isolate the benefits of our dataset and design beyond naïve fusion strategies.

To further assess on-device deployment feasibility, we evaluated the \textit{8-bit quantized} version of our fine-tuned models. This approach reduces memory by up to fourfold, lowering computational demands while maintaining competitive performance. Though quantization may slightly reduce accuracy, the simplicity of the rewriting task makes this trade-off worthwhile. We examine whether an 8-bit model can efficiently handle multimodal instruction rewriting while staying lightweight for real-world use.

\begin{table*}[t]
    \centering
    \footnotesize
    \begin{tabular}{lccccc|cc}
        \toprule    
        \textbf{Model} & \multicolumn{2}{c}{\textbf{ROUGE-N}} & \textbf{ROUGE-L} & \textbf{BLEU} & \textbf{MET-} & \textbf{Intent} & \textbf{Arg} \\
                       & N=1              & N=2              &                  &               & \textbf{EOR}  & \textbf{Acc}   & \textbf{MJS} \\
        \midrule
        TextOnly\textsubscript{1a}: Qwen2.5-0.5B\textsubscript{EasyOCR+Meta} & 19.6 & 7.3 & 18.2 & 1.8 & 24.5 & 45.0 & 47.8  \\
        TextOnly\textsubscript{1b}: MistralLite-150M\textsubscript{EasyOCR+Meta} & 7.1 & 0.9 & 6.2 & 0.2 & 12.1 & 24.9 & 45.1 \\ 
        \midrule
        BL\textsubscript{1a}: PaliGemma2-10B\textsubscript{vanilla} & 3.4 & 0.5 & 3.3 & 0.03 & 2.3 & 16.2 & 42.7 \\
        BL\textsubscript{1b}: Qwen2-VL-7B\textsubscript{vanilla}   & 43.7 & 24.7 & 40.8 & 12.3 & 39.5 & 50.3 & 65.2 \\
        BL\textsubscript{2a}: PaliGemma2-10B\textsubscript{Self\_Caption+EasyOCR} & 11.1 & 2.5 & 11.1 & 0.03 & 4.5 & 19.3 & 30.0 \\
        BL\textsubscript{2b}: Qwen2-VL-7B\textsubscript{Self\_Caption+EasyOCR} & 41.3 & 24.0 & 38.7 & 8.4 & 39.1 & 61.2 & 67.0 \\
        \midrule
        BL\textsubscript{3a}: SmolVLM\textsubscript{FT} & 35.8 & 19.6 & 33.5 & 7.9 & 40.1 & 49.7 & 59.5 \\
        BL\textsubscript{3b}: SmolVLM\textsubscript{Metadata+EasyOCR} & 23.3 & 10.2 & 21.2 & 3.2 & 26.2 & 21.5 & 49.2 \\
        BL\textsubscript{3c}: SmolVLM\textsubscript{Self\_Caption+EasyOCR} & 17.2 & 6.4 & 15.8 & 2.2 & 17.1 & 24.1 & 47.2 \\
        \midrule
        ReVision-BL                         & 56.9 & 41.4 & 55.4 & 27.7 & 61.4 & 56.5 & 68.8 \\
        ReVision-Metadata\textsubscript{GPT-4o\_Caption+OCR} & 72.4 & 60.6 & 71.5 & 49.9 & 74.4 & 62.4 & 73.7 \\
        ReVision-Metadata\textsubscript{Self\_Caption+EasyOCR}     & \textbf{79.3} & \textbf{70.0} & \textbf{78.4} & \textbf{61.5} & \textbf{80.2} & \textbf{71.5} & 74.5 \\
                ReVision-Metadata\textsubscript{Self\_Caption+EasyOCR(8bit)} & \textit{79.2} & \textit{69.9} & \textit{78.3} & \textit{61.3} & \textit{80.1} & \textit{67.6} & \textbf{\textit{79.5}} \\

        \bottomrule
    \end{tabular}
    \caption{Evaluation Results for Baseline and RV Models as a Percentage. BL = Baseline; ROUGE-N = N-grams between the system and reference summaries; ROUGE-L = Longest common subsequence-based statistics; BLEU = BiLingual Evaluation Understudy; METEOR = Metric for Evaluating Translation with Explicit Ordering; Intent Acc = Intent Accuracy; Arg MJS = Argument Mean Jaccard Similarity.}

    \label{tab:evaluation_results}
\end{table*}
\section{Evaluation Metrics}
To evaluate our models in Visual Instruction Rewriting, we use standard NLG metrics (BLEU, METEOR, ROUGE) \cite{sharma2017nlgeval} alongside task-specific semantic parsing evaluations. While NLG metrics assess linguistic similarity, they do not capture functional quality in downstream AI systems. \textbf{Effective rewriting must also ensure instructions remain interpretable by semantic parsers extracting intent and arguments} \cite{louvan2020recent}. In the absence of an existing parser tailored to our ontology (Figure~\ref{fig:intent_argument_box}), we employ GPT-4 as a proxy to simulate an on-device parser for intent classification and argument extraction. To evaluate intent and structure preservation, we compare GPT-4-generated parses for both reference and model-generated rewrites. For clarity, we present a collapsed view of intents and arguments. The following metrics are used for evaluation

\begin{itemize}
    \item \textbf{Intent Accuracy:} Exact match of intent labels between reference and model-generated rewrites, assessing task-specific intent preservation. 
    
    \item \textbf{Argument Similarity:} Mean Jaccard Similarity (MJS) between argument labels from reference and model rewrites, ensuring retention of key task-related arguments.
\end{itemize}
\begin{figure*}[t]
  \includegraphics[width=\textwidth]{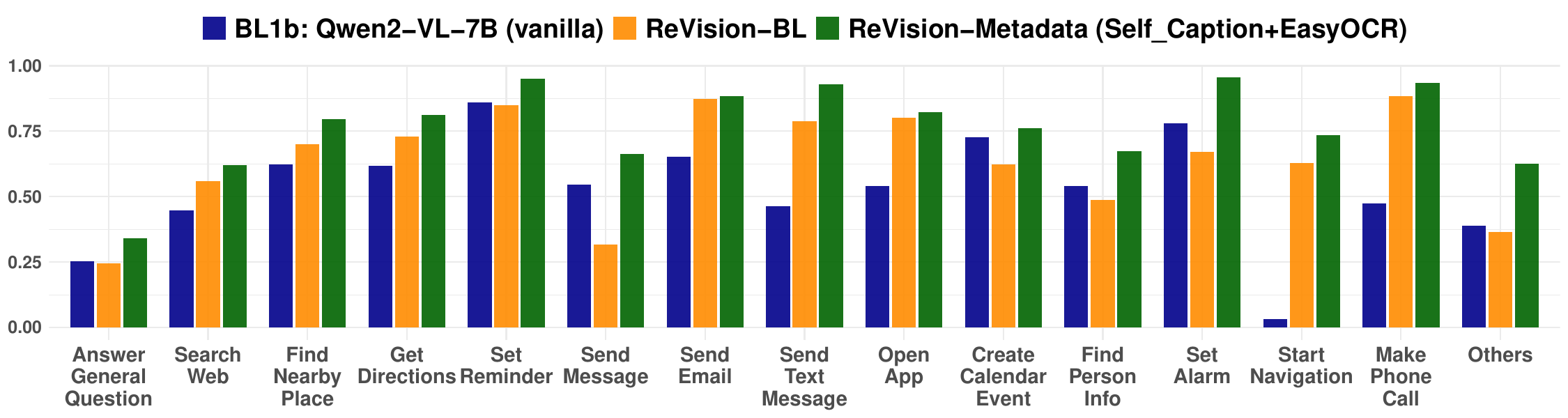}
  \caption{Class-wise F1 Scores for Intent Classification}
  \label{fig:F1}
\end{figure*}
\section{Results and Discussion}
\label{sec:results}

Table \ref{tab:evaluation_results} presents the evaluation results for both baseline models (BL) and our proposed ReVision models across Language Generation (NLG) metrics (ROUGE, BLEU, METEOR) and semantic parsing performance (Intent Accuracy and Argument Mean Jaccard Similarity). We also provide anecdotal examples in Figure \ref{fig:anecdotal_examples} to illustrate the strengths and limitations of various models.

Both \textbf{TextOnly} baseline variants performed significantly worse than ReVision, highlighting the value of multimodal inputs. These models struggled with proper nouns and named entities from captions and OCR, showing high sensitivity to metadata quality. Midsize baseline VLMs underperformed not due to weak modeling but due to lack of tuning for rewriting. Though \textit{PaliGemma2-10B} and \textit{QwenVL-7B} perform well on vision-language tasks, they are not optimized for meta-instruction following, as seen in their vanilla versions (BL\textsubscript{1a}, BL\textsubscript{1b}) with low ROUGE-1 (3.4\%, 43.7\%), negligible BLEU (0.03\%, 12.3\%), and poor Intent Accuracy (16.2\%, 50.3\%). They often misinterpret rewriting as direct response or autocompletion, especially with imperative inputs, leading to refusal (``I can't help with that'') or incorrect completions---hurting NLG and parsing metrics. Their small size (<10B parameters) limits instruction-following and world knowledge needed for structured rewriting. Adding \textit{Self\_Caption+EasyOCR} metadata (BL\textsubscript{2a}, BL\textsubscript{2b}) slightly helps, notably for \textit{QwenVL-7B} (Intent Accuracy: 50.3\%~$\rightarrow$~61.2\%), but ROUGE and BLEU remain low, showing the need for instruction tuning. The fine-tuned \textit{SmolVLM-256M} baseline also underperforms with default tuning, often over-explaining by adding unnecessary descriptions and artifacts, likely due to pretraining on detailed tasks (e.g., video narration). While suboptimal here, \textit{SmolVLM} remains a promising candidate with targeted tuning and prompting.


In contrast, our proposed \textsc{ReVision} models, explicitly trained for rewriting, substantially outperform all baselines, demonstrating the importance of task-specific tuning. Even without metadata, \textsc{ReVision-BL} exceeds input-augmented baselines with ROUGE-1 of 56.9\%, BLEU of 27.7\%, and Intent Accuracy of 56.5\%, highlighting that a compact, instruction-tuned VLM can surpass larger, non-specialized models---an observation further supported by the intent category-wise F1 scores in Figure~\ref{fig:F1}. Incorporating metadata yields additional gains: \textsc{ReVision-Metadata}, enhanced with GPT-4-derived captions and OCR text, achieves 72.4\% ROUGE-1, 49.9\% BLEU, and 62.4\% Intent Accuracy, confirming that extracted text aids in resolving multimodal ambiguities. The top-performing model, \textsc{ReVision-Metadata-Self\_Caption+EasyOCR}, achieves the highest scores across all metrics, showing that even lightweight captioning and OCR tools can enhance rewriting quality. Furthermore, the 8-bit quantized version of this model delivers nearly equivalent performance to its full-precision counterpart---67.6\% vs.\ 71.5\% Intent Accuracy---while slightly improving Argument Similarity (79.5\%), indicating its suitability for deployment on resource-constrained devices.

Despite the strong performance of our \textit{ReVision} variants, certain limitations hinder further accuracy gains. A primary issue is the loss of fine-grained text details caused by downsampling images to $256 \times 256$ resolution, which impairs recognition of critical elements such as ingredient lists or nutritional facts on product packaging. Additionally, the dataset's lack of explicit reference localization limits the model's ability to align user intent with specific image regions, resulting in object disambiguation and instruction alignment errors. Future work could address these challenges by incorporating bounding box annotations to provide spatial grounding cues and by processing localized image regions rather than entire downsampled images, reducing information loss in text-heavy visual inputs. This approach aligns with Pali-Gemma’s short-resolution increase technique~\cite{beyer2024paligemma}, which improves fine-grained visual understanding. Nonetheless, our findings reaffirm that task-specific instruction tuning and metadata augmentation markedly enhance multimodal rewriting, supporting scalable and efficient on-device deployment.

\section{Conclusion and Future Work}
\label{sec:conclusion}
In this work, we explored Visual Instruction Rewriting as a lightweight, privacy-preserving approach to multimodal interaction on AR, VR, and smartphone devices. With a strong emphasis on dataset development, we present a diverse collection of 39,000+ examples covering 14 domains, enabling robust training for on-device instruction rewriting. Our approach ensures that text-only inference is more secure in privacy-sensitive settings by \textbf{eliminating the need to send personal vision-related images to the server}, reducing data exposure risks. Additionally, rewriting removes the necessity of storing images, making multimodal AI systems more efficient and privacy-focused. Our experimental results show that even an 8-bit quantized model maintains strong performance while significantly reducing memory requirements. For future work, we aim to expand data coverage by incorporating more diverse real-world multimodal instructions and introducing multilingual support to enhance accessibility. Furthermore, improving deixis resolution with bounding box annotations and localized image region training will enhance reference grounding, while integrating gaze tracking and tactile input can further refine contextual understanding in on-device AI assistants.

\section*{Limitations}
While our approach demonstrates strong performance in Visual Instruction Rewriting, several limitations remain. First, image downsampling to $256 \times 256$ resolution can lead to the loss of fine-grained text details, affecting instructions that rely on small-font information, such as nutritional labels or product specifications. Second, deictic reference resolution remains challenging, especially in images with multiple similar objects where the model lacks explicit localization cues. The absence of bounding box annotations in our dataset limits the model’s ability to disambiguate references, leading to errors in object-grounded instructions. Additionally, while our model is lightweight and optimized for on-device execution, it still lags behind larger VLMs in handling complex multimodal instructions requiring deep reasoning and external world knowledge. Lastly, our dataset, while diverse across 15+ domains, is monolingual, limiting applicability to multilingual and culturally varied settings. Future work can address these challenges by increasing dataset coverage, incorporating localized image region processing, and adding bounding box annotations to improve reference resolution and multimodal grounding.

\section*{Ethical Considerations}
This work prioritizes privacy and ethical considerations by designing a lightweight, on-device Visual Instruction Rewriting system that eliminates the need to transmit personal vision-related data to external servers. By converting multimodal instructions into text-only commands, our approach reduces data exposure risks and ensures secure, user-controlled inference. Our dataset is sourced from publicly available and academic-use image collections, ensuring compliance with fair use and licensing policies. However, we acknowledge potential biases in data distribution and the need for greater multilingual and cultural inclusivity. Future efforts will focus on expanding dataset diversity, improving fairness in multimodal understanding, and ensuring responsible AI deployment in real-world applications.

Additionally, we acknowledge the use of OpenAI’s ChatGPT-4 system solely for enhancing writing efficiency, generating LaTeX code, and aiding in error debugging. No content related to the survey's research findings, citations, or factual discussions was autogenerated or retrieved using Generative AI-based search mechanisms. Our work remains grounded in peer-reviewed literature and ethical academic standards.

\bibliography{acl2025}
\bibliographystyle{acl_natbib}

\appendix
\section{Appendix}
\label{sec:appendix}

\begin{figure*}[t]
\centering
\begin{tcolorbox}[colback=gray!10, colframe=gray!30, coltitle=black,
    fonttitle=\bfseries, 
    title=Intent and Argument Labels]
\textbf{Intent Labels:}  
    AdjustBrightness, AdjustTemperature, AdjustVolume, AnswerGeneralQuestion, CheckSecurityCamera, CheckStockPrice, CheckTraffic, CheckVoicemail, CheckWeather, ConvertUnits, CreateCalendarEvent, DefineWord, EstimateArrivalTime, FindNearbyPlace, FindPersonInfo, GetDirections, GetFact, GetNewsUpdate, GetSportsScores, LockDoor, MakeCall, MakePhoneCall, MathCalculation, OpenApp, PauseMusic, PlayMusic, PlayPodcast, PlayVideo, ReadMessage, ReplyToMessage, SearchMovie, SearchWeb, SendEmail, SendGroupMessage, SendMessage, SendTextMessage, SetAlarm, SetPlaybackSpeed, SetReminder, SetScene, SetTimer, ShowTVGuide, SkipTrack, StartNavigation, StartVacuum, StartVideoCall, StopNavigation, StopVacuum, TranslateText, TurnOffDevice, TurnOnDevice, UnlockDoor 
    
    \vspace{5pt}
    
    \textbf{Argument Labels:}  
    AlarmTime, AppName, ArtistName, BrightnessLevel, CameraLocation, ContactName, CurrentLocation, DateTime, Destination, DeviceName, ETA, EmailBody, EmailSubject, EpisodeTitle, EventDateTime, EventLocation, EventTitle, LanguagePair, LockState, MathExpression, MessageBody, MessageContent, MovieName, NewsTopic, PersonName, PlaceCategory, PlaybackSpeed, PodcastTitle, QueryText, QuestionText, Recipient, RecipientName, ReminderContent, RouteType, SceneName, SongName, SportEvent, StockSymbol, TVChannel, TemperatureValue, TimerDuration, UnitToConvert, VoicemailSender, VolumeLevel, WeatherDate, WeatherLocation, WordToDefine 
    \end{tcolorbox}
    \caption{Intent and Argument Labels Considered for Data Bootstrapping}
    \label{fig:intent_argument_box}
\end{figure*}

\begin{figure*}[t]
    \centering
    \begin{tcolorbox}[colback=gray!10, colframe=gray!30, coltitle=black,
        fonttitle=\bfseries, 
        title=Intent and Argument Labels]
        
    \textbf{Intent Labels:}  
    AnswerGeneralQuestion, CreateCalendarEvent, FindNearbyPlace, FindPersonInfo, GetDirections, MakePhoneCall, OpenApp, SearchWeb, SendEmail, SendMessage, SendTextMessage, SetAlarm, SetReminder, StartNavigation, Others
    
    \vspace{5pt}
    
    \textbf{Argument Labels:}  
    AlarmTime, AppName, ArtistName, BrightnessLevel, CameraLocation, ContactName, CurrentLocation, DateTime, DeviceName, ETA, EmailBody, EpisodeTitle, EventTitle, LanguagePair, LockState, MathExpression, MovieName, NewsTopic, PlaceCategory, PlaybackSpeed, PodcastTitle, QueryText, ReminderContent, RouteType, SceneName, SongName, SportEvent, StockSymbol, TVChannel, TemperatureValue, UnitToConvert, VoicemailSender, VolumeLevel
    \end{tcolorbox}
    \caption{Collapsed Intent and Argument Labels for Metric Computation}
    \label{fig:intent_argument_collapsed_box}
\end{figure*}
\begin{figure*}[t]
  \centering
  \includegraphics[width=0.95\textwidth]{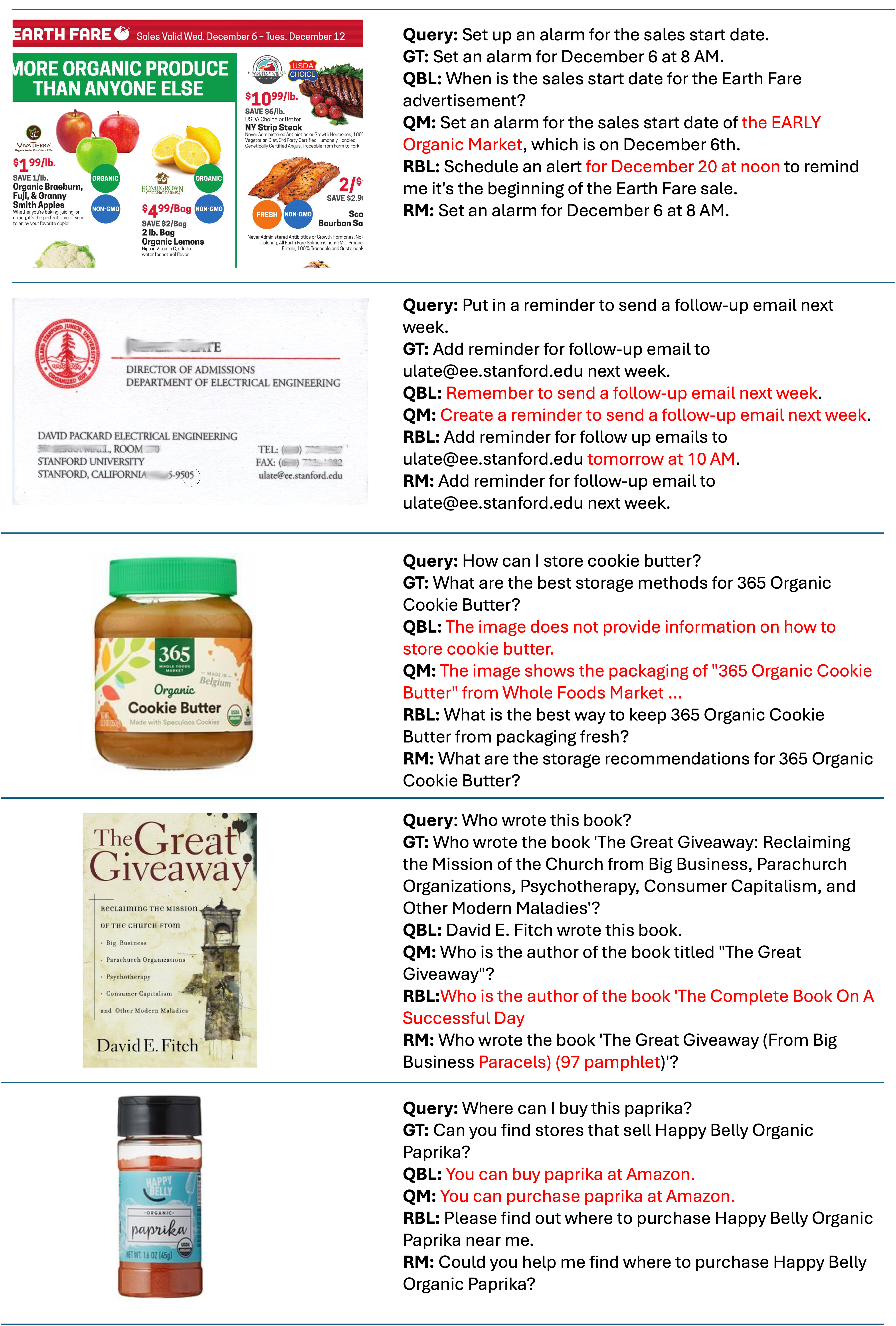} 
    \caption{Anecdotal examples illustrating images, queries, and rewrites across different domains. Abbreviations: \textbf{GT} $\rightarrow$ Ground Truth, \textbf{QBL} $\rightarrow$ Qwen Baseline, \textbf{QM} $\rightarrow$ Qwen with Self-Caption and EasyOCR Metadata, \textbf{RBL} $\rightarrow$ ReVision (ours) Baseline, \textbf{RM} $\rightarrow$ ReVision (ours) with Self-Caption and EasyOCR Metadata. Incorrect and hallucinatory output phrases are highlighted in red.}

  \label{fig:anecdotal_examples}
\end{figure*}

\end{document}